\documentclass[conference]{IEEEtran}
\IEEEoverridecommandlockouts
\usepackage{amsmath,amssymb,amsfonts}
\usepackage{algorithmic}
\usepackage[disable,pdf]{epstopdf}
\usepackage{graphicx}
\usepackage{textcomp}

\usepackage{multirow}
\usepackage{booktabs}

\usepackage{biblatex}
\usepackage{xcolor}
\usepackage{hyperref}

\usepackage{geometry}

\def\BibTeX{{\rm B\kern-.05em{\sc i\kern-.025em b}\kern-.08em
    T\kern-.1667em\lower.7ex\hbox{E}\kern-.125emX}}
    
\begin{document}

\title{VisuoAlign: Safety Alignment of LVLMs with Multimodal Tree Search}

\author{
\IEEEauthorblockN{
MingSheng Li$^{1*}$, 
Guangze Zhao$^{2*}$, 
Sichen Liu$^{3\dagger}$
}
\IEEEauthorblockA{
$^{1}$Independent Researcher, \texttt{daroubao1425@gmail.com} \\
$^{2}$Harbin Institute of Technology \texttt{zhaoguangze@chanct.com} \\
$^{3}$Xi’an Jiaotong-Liverpool University, \texttt{Sichen.Liu@xjtlu.edu.cn}
}
\thanks{$^{*}$Equal contribution. $^{\dagger}$Corresponding author. This work is supported by National Natural Science Foundation of China
(Grant No. 12404538), State Key Laboratory of Acoustics and Marine In-
formation, Chinese Academy of Sciences (Grant No. SKLA202413) and
Xi’an Jiaotong-Liverpool University, Research Development Fund (Grant
No. RDF-22-02-029).}
}

\maketitle

\begin{abstract}

Large Vision-Language Models (LVLMs) have achieved remarkable progress in multimodal perception and generation, yet their safety alignment remains a critical challenge. Existing defenses are vulnerable to multimodal jailbreaks, as visual inputs introduce new attack surfaces, reasoning chains lack safety supervision, and alignment often degrades under modality fusion. To overcome these limitations, we propose VisuoAlign, a framework for multi-modal safety alignment via prompt-guided tree search. VisuoAlign embeds safety constraints into the reasoning process through visual–textual interactive prompts, employs Monte Carlo Tree Search (MCTS) to systematically construct diverse safety-critical prompt trajectories, and introduces prompt-based scaling to ensure real-time risk detection and compliant responses. Extensive experiments demonstrate that VisuoAlign proactively exposes risks, enables comprehensive dataset generation, and significantly improves the robustness of LVLMs against complex cross-modal threats.

\end{abstract}


\renewcommand{\thefootnote}{} 

\renewcommand{\thefootnote}{\arabic{footnote}}

\section{Introduction}
Large Vision-Language Models (LVLMs)\cite{li2025clipscore} have recently demonstrated impressive progress in perception and generation across diverse multimodal tasks\cite{li2024tuni,cheng2025gibberish,duan2025oyster,li2025faedkv,long2025truthful,fu2025pruning,fu2025quantized}. However, ensuring safety alignment in such models remains a pressing challenge. Despite their remarkable capabilities, LVLMs are prone to jailbreak attacks\cite{teng2024heuristic} in cross-modal interactions, often underestimating potential risks or producing unsafe outputs in complex scenarios \cite{qi2024visual,fang2025safemlrm}.

Existing safety alignment methods \cite{cheng2024reinforcement,cheng2025llm,cheng2025inverse} exhibit several limitations. First, visual inputs introduce new attack surfaces: hidden prompt injections embedded in images or adversarial visual perturbations can bypass traditional text-based filters, exposing LVLMs to multimodal jailbreaks \cite{zhao2025strata,cheng2024pbi}. Second, current defense strategies predominantly focus on final outputs while neglecting the intermediate reasoning process in visual-textual interactions, leaving critical vulnerabilities in the reasoning chain unmonitored \cite{yeh2025adversarial,zhou2024role}. Third, modality fusion frequently causes alignment degradation, as safety constraints learned in text-only spaces fail to transfer effectively to multimodal contexts \cite{liu2024unraveling,wang2025safety}. Overall, existing approaches rely heavily on static refusals or handcrafted rules, which lack adaptability to evolving, diverse, and complex cross-modal threat scenarios.

To address these challenges, we propose \textbf{VisuoAlign}, a novel framework for multi-modal safety alignment via prompt-guided tree search. Unlike prior approaches, VisuoAlign embeds safety constraints directly into the reasoning process by interleaving visual–textual interactive prompts, systematically explores diverse prompt trajectories with Monte Carlo Tree Search (MCTS)~\cite{browne2012survey} to construct a broad and adversarially aware alignment dataset, and introduces prompt-based safety scaling during inference through multi-step safe prompts and dynamic refusal or justification nodes. This design enables proactive risk exposure, comprehensive dataset generation, and real-time safety adaptation, thereby enhancing the robustness of LVLMs against complex cross-modal threats. Extensive experiments on various benchmarks demonstrate that VisuoAlign substantially reduces jailbreak success rates while achieving alignment safety rates above 0.92, outperforming all baseline methods.

In summary, our contributions are as follows:
\begin{itemize}
\item We embed safety constraints into multimodal prompts and interleave them within reasoning, achieving intrinsic safety awareness in visual–textual interaction.
\item We leverage MCTS to construct a broad, adversarially aware safety alignment dataset and introduce prompt-based safety scaling at inference time for real-time risk adaptation.
\item Extensive evaluations on four safety benchmarks  demonstrate that VisuoAlign achieves superior robustness, strong cross-model generalization, and practical efficiency over existing baselines.
\end{itemize}

\begin{figure*}[t]
\centering
\includegraphics[width=\linewidth]{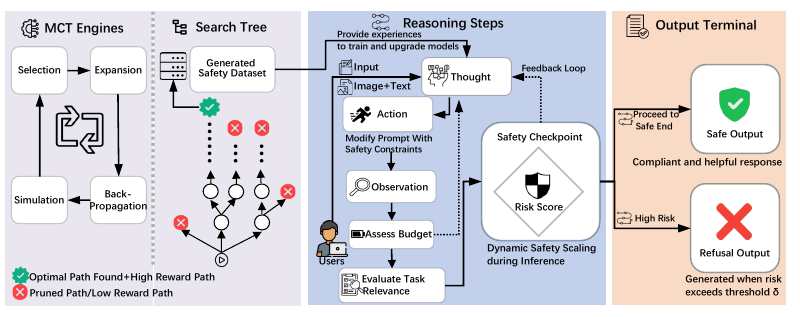}
\caption{Overview of VisuoAlign.\label{VisuoAlign}}
\end{figure*}

\section{Related Work}

\paragraph{Safety Alignment.}  

Safety alignment is a key challenge in developing large language models (LLMs). Existing methods such as reinforcement learning from human feedback (RLHF) \cite{ouyang2022training,liu2024upper,hsu2024randomized,hesample}, direct preference optimization (DPO) \cite{rafailov2023direct}, and self-alignment  have greatly reduced unsafe generations \cite{cao2025agr,}. Yet, research on \textbf{multi-modal safety alignment} is limited. Extending alignment to large vision-language models (LVLMs) raises unique issues, including visual prompt injection \cite{clusmann2025prompt}, cross-modal jailbreaks \cite{qi2024visual,jiang2025cross}, and weakened robustness after modality fusion \cite{fang2025safemlrm}. Current solutions, often based on static refusal or rule-based filters \cite{pollock1988rule}, lack adaptability and fail to provide fine-grained safety regulation during reasoning.

\paragraph{Tree Search.}  
Tree search, especially Monte Carlo Tree Search (MCTS) \cite{browne2012survey}, has been widely used for planning and decision-making\cite{liao2025survey,li2025medguide}. Recent work applies it to LLM reasoning \cite{yao2023tree}, enabling test-time exploration of reasoning paths. While effective in text-based tasks, its role in multimodal safety alignment remains underexplored. Combining tree search with multimodal prompts can generate diverse safety-critical trajectories, simulate risks, and dynamically guide LVLMs toward safe and robust outputs.

\section{Methodology}

We propose \textbf{VisuoAlign}, a novel framework for multi-modal safety alignment that embeds safety constraints into the reasoning process via prompt-guided tree search as shown in Fig \ref{VisuoAlign}. Unlike prior approaches that rely on static filtering or refusal policies, VisuoAlign dynamically integrates visual--textual interactive prompts and explores diverse reasoning trajectories with Monte Carlo Tree Search (MCTS), enabling both proactive risk detection and adaptive safety scaling during inference.  

\subsection{Safety-Constrained Multimodal Reasoning}

Given an input multimodal query $Q = \{I, T\}$
where $I$ denotes the visual input (e.g., an image) and $T$ the textual input, our framework produces a sequence of reasoning steps $A = \{a_1, a_2, \ldots, a_{|A|}\}$
and finally provides a safe output response $R$.  

Each reasoning step $a_t$ is decomposed into three phases:  
(1) \textit{Thought}: the LVLM processes both $I$ and $T$, generating candidate reasoning directions while estimating potential safety risks.  
(2) \textit{Action}: a safety-aware prompt modification or expansion is applied.  
(3) \textit{Observation}: the updated multimodal context is re-evaluated for the next step.  

The process is constrained by a safety-aware trajectory distribution:
\begin{align*}
    A \sim P_\theta(A \mid Q), \quad \text{s.t. } \sum_{t=1}^{|A|} \mathcal{C}(a_t) \leq \tau,
\end{align*}
where $\mathcal{C}(a_t)$ denotes the safety complexity cost of each step, and $\tau$ is the maximum safety budget.

\subsection{Monte Carlo Tree Search for Safe Data Construction}

We model the prompt design space as a search tree $\mathcal{T}$, where state node $s_t$ corresponds to the current multimodal reasoning context $(I, T, A_{1:t})$, action node represents a prompt modification (e.g., inserting a safety checkpoint) and reward $r_t$ is computed as follows:
\begin{align*}
    r_t = \alpha \cdot \text{Safe}(s_t) + \beta \cdot \text{Comp}(s_t),
\end{align*}
where $\text{Safe}(\cdot)$ measures alignment and $\text{Comp}(\cdot)$ measures task fidelity \cite{simon2009balancing}, $\alpha$ and $\beta$ are weight coefficients.

MCTS operates in the following four phases:  

\begin{enumerate}
    \item \textbf{Selection:}  
    \begin{align*}
        s^\ast = \arg\max_{s \in \mathcal{T}} \left[ \bar{r}(s) + c \sqrt{\frac{\ln N}{n(s)}} \right],
    \end{align*}
    where $\bar{r}(s)$ is the average reward, $n(s)$ the visit count, $N$ the total visits, and $c$ is a trade-off parameter.

    \item \textbf{Expansion:} Generate candidate prompts $\{s_1^t, \ldots, s_k^t\}$.  

    \item \textbf{Simulation:} Extend each candidate until termination, producing outcomes $r_i^t$.  

    \item \textbf{Backpropagation:} Update rewards along the trajectory.  

\end{enumerate}

Finally, the dataset is formalized as:
{\small
\begin{align*}
\textstyle
    \mathcal{D}_{\text{safe}} = \bigcup_{Q \in \mathcal{Q}} \{ (Q, A, R) \mid A \sim \text{MCTS}(Q), R \in \text{SafeOutputs} \}.
\end{align*}
}

\subsection{Dynamic Safety Scaling during Inference}

During inference, we enhance safety through \textit{prompt-based scaling}, which integrates explicit safety checks into the reasoning process. For instance, the model can be guided by the following structured prompt:

\begin{quote}
For each candidate reasoning step:  
(1) Evaluate semantic relevance to the task.  
(2) Compute a safety risk score in $[0,1]$.  
(3) If the risk score exceeds $0.5$, prune this reasoning path and output a refusal response.  
(4) Otherwise, proceed with the candidate path that minimizes risk while maintaining task fidelity.  
\end{quote}

This design ensures that safety constraints are continuously enforced throughout the reasoning trajectory, allowing the model to adaptively filter unsafe paths and prioritize compliant outputs.

At each step, we greedily select action as follows:
\begin{align*}
    a_t^\ast = \arg\max_{a_t \in \mathcal{A}} \Big[ \gamma \cdot P_\theta(a_t \mid s_{t-1}) - \lambda \cdot \text{Risk}(a_t) \Big],
\end{align*}
where $\gamma$ controls task fidelity and $\lambda$ penalizes unsafe generations.  

Dynamic refusal and justification nodes are injected when $\text{Risk}(a_t) > \delta$, where $\delta$ is a threshold. The final safe output is:
\begin{align*}
    R = f(Q) = \text{Scale}\big( \text{MCTS}(Q), \delta, \lambda \big).
\end{align*}

By embedding safety checkpoints within multimodal prompts, systematically constructing datasets with MCTS, and applying safety scaling during inference, \textbf{VisuoAlign} transforms safety alignment from a static filter into a dynamic reasoning-integrated mechanism, improving robustness against multimodal jailbreaks.
\section{Experiments}

\subsection{Experimental Setup}

\paragraph{Models}

We conduct our evaluation on three widely-used open-source LVLMs: LLaVA-1.5-7B~\cite{liu2023visual}, MiniGPT-4-7B~\cite{zhu2023minigpt}, and Qwen-VL-7B~\cite{bai2023qwen}. 

\paragraph{Baselines}

To comprehensively validate the efficacy of VisuoAlign, we compare it against five categories of baselines: (1) \textbf{Base}: The original models without any safety alignment; (2) \textbf{RLHF}~\cite{ouyang2022training}: Alignment through reinforcement learning from human feedback; (3) \textbf{DPO}~\cite{rafailov2023direct}: Direct preference optimization; (4) \textbf{StaticRefusal}~\cite{pollock1988rule}: A baseline employing a static, rule-based refusal policy; and (5) \textbf{SEA}~\cite{zhang2025spa}, a state-of-the-art method for multimodal safety alignment.

\paragraph{Implementation Details}
We use the PyTorch framework and fine-tune models with LoRA(rank 16, $\alpha=32$) using the AdamW optimizer (learning rate $1\times10^{-4}$, batch size 4). Our alignment dataset, $\mathcal{D}_{\text{safe}}$, comprises 10K trajectories generated by running MCTS on prompts from HarmfulQA~\cite{zhou2023making} augmented with images from LAION~\cite{schuhmann2022laion}. All experiments were conducted on four NVIDIA A100 (80GB) GPUs.

\paragraph{Benchmarks and Metrics}
Evaluation is performed on four standard multimodal safety benchmarks: \textbf{MultiHarm}~\cite{qi2024visual}, \textbf{V-Jailbreak}~\cite{cheng2024pbi}, \textbf{MM-SafetyBench}~\cite{liu2024mm}, and \textbf{SPA-VL}~\cite{zhang2025spa}. The key metrics are: \textbf{JSR} (Jailbreak Success Rate, $\downarrow$), \textbf{ASR} (Alignment Safety Rate, $\uparrow$), and \textbf{AR} (Attack Robustness, $\downarrow$).

\subsection{Overall Performance}

We evaluate VisuoAlign on the core MultiHarm and V-Jailbreak benchmarks (Table~\ref{tab:main_results}). VisuoAlign consistently outperforms all baselines across models, achieving new state-of-the-art results.

Compared to base models, it reduces JSR by over 57\% on average and raises ASR above 0.92. On the challenging V-Jailbreak dataset, where RLHF and DPO degrade, VisuoAlign maintains low JSR (0.28 on LLaVA) and high ASR (0.93), showing strong robustness against cross-modal threats. Similar gains are observed on MM-SafetyBench and SPA-VL.

\begin{table}[h]
\centering
\caption{Main results on MultiHarm and V-Jailbreak. VisuoAlign achieves the lowest JSR and highest ASR across all models.}
\label{tab:main_results}
\resizebox{\columnwidth}{!}{%
\begin{tabular}{l l cc cc}
\toprule
\multirow{2}{*}{\textbf{Model}} & \multirow{2}{*}{\textbf{Method}} & \multicolumn{2}{c}{\textbf{MultiHarm}} & \multicolumn{2}{c}{\textbf{V-Jailbreak}} \\
\cmidrule(lr){3-4} \cmidrule(lr){5-6}
& & JSR$\downarrow$ & ASR$\uparrow$ & JSR$\downarrow$ & ASR$\uparrow$ \\
\midrule
\multirow{6}{*}{LLaVA-1.5-7B}
& Base & 0.78 & 0.22 & 0.81 & 0.19 \\
& RLHF & 0.62 & 0.48 & 0.65 & 0.44 \\
& DPO & 0.59 & 0.51 & 0.63 & 0.47 \\
& StaticRefusal & 0.55 & 0.65 & 0.60 & 0.59 \\
& SEA & 0.48 & 0.72 & 0.53 & 0.66 \\
& \textbf{VisuoAlign} & \textbf{0.33} & \textbf{0.92} & \textbf{0.28} & \textbf{0.93} \\
\midrule
\multirow{6}{*}{Qwen-VL-7B}
& Base & 0.75 & 0.25 & 0.79 & 0.21 \\
& RLHF & 0.58 & 0.52 & 0.61 & 0.49 \\
& DPO & 0.56 & 0.54 & 0.59 & 0.51 \\
& StaticRefusal & 0.52 & 0.68 & 0.56 & 0.62 \\
& SEA & 0.45 & 0.75 & 0.50 & 0.69 \\
& \textbf{VisuoAlign} & \textbf{0.30} & \textbf{0.95} & \textbf{0.26} & \textbf{0.96} \\
\bottomrule
\end{tabular}%
}
\end{table}

\subsection{In-depth Analysis}

\paragraph{Ablation Study}

We perform ablation studies on LLaVA-1.5-7B (Table~\ref{tab:ablation}) to assess each component of VisuoAlign. Removing MCTS increases JSR by 22\%, disabling dynamic scaling reduces ASR by 18\%, and omitting safety constraints degrades both metrics. These results confirm that embedding safety into reasoning is central to VisuoAlign’s effectiveness.

\begin{table}[h]
\centering
\caption{Ablation study on LLaVA-1.5-7B on the MultiHarm benchmark. Each component is critical for performance.}
\label{tab:ablation}
\resizebox{0.9\columnwidth}{!}{%
\begin{tabular}{lcc}
\toprule
\textbf{Variant} & \textbf{JSR}$\downarrow$ & \textbf{ASR}$\uparrow$ \\
\midrule
\textbf{VisuoAlign (Full)} & \textbf{0.33} & \textbf{0.92} \\
\midrule
w/o MCTS (Random Sampling) & 0.55 & 0.74 \\
w/o Dynamic Scaling & 0.41 & 0.74 \\
w/o Safety Constraints & 0.48 & 0.68 \\
\bottomrule
\end{tabular}%
}
\end{table}

\paragraph{Robustness to Advanced Attacks}

We evaluate VisuoAlign against Visual Prompt Injection (VPI)~\cite{cheng2024pbi}, Adversarial Perturbations (AP)~\cite{madry2017towards}, and Cross-Modal Fusion (CMF)~\cite{qi2024visual}. As shown in Table~\ref{tab:robustness}, it cuts attack success rates by 65--78\% over the base model, thanks to its tree-search mechanism that proactively prunes unsafe reasoning paths—unlike conventional fine-tuning methods.

\begin{table}[h]
\centering
\caption{Robustness (AR$\downarrow$) against advanced attack types on LLaVA-1.5-7B. Lower is better.}
\label{tab:robustness}
\resizebox{0.6\columnwidth}{!}{%
\begin{tabular}{l ccc}
\toprule
\textbf{Method} & \textbf{VPI} & \textbf{AP} & \textbf{CMF} \\
\midrule
Base & 0.85 & 0.82 & 0.79 \\
RLHF & 0.70 & 0.68 & 0.65 \\
DPO & 0.67 & 0.65 & 0.62 \\
StaticRefusal & 0.62 & 0.60 & 0.58 \\
SEA & 0.55 & 0.72 & 0.52 \\
\textbf{VisuoAlign} & \textbf{0.22} & \textbf{0.18} & \textbf{0.25} \\
\bottomrule
\end{tabular}%
}
\end{table}

\paragraph{Sensitivity Analysis}

We analyze sensitivity to the inference hyperparameters—the risk penalty $\lambda$ and refusal threshold $\delta$ (Table~\ref{tab:sensitivity}). Varying one while fixing the other at its optimum ($\lambda=0.6, \delta=0.5$), we find VisuoAlign remains stable: JSR and ASR stay near-optimal for $\lambda \in [0.4,0.8]$. This low sensitivity shows VisuoAlign is robust and easy to deploy without fine-grained tuning.

\begin{table}[h]
\centering
\caption{Sensitivity analysis on LLaVA-1.5-7B (MultiHarm benchmark). The optimal setting is bolded.}
\label{tab:sensitivity}
\resizebox{0.9\columnwidth}{!}{%
\begin{tabular}{lcc | lcc}
\toprule
\multicolumn{3}{c|}{\textbf{Varying $\lambda$ (fixed $\delta=0.5$)}} & \multicolumn{3}{c}{\textbf{Varying $\delta$ (fixed $\lambda=0.6$)}} \\
\cmidrule(r){1-3} \cmidrule(l){4-6}
$\lambda$ & JSR$\downarrow$ & ASR$\uparrow$ & $\delta$ & JSR$\downarrow$ & ASR$\uparrow$ \\
\midrule
0.2 & 0.45 & 0.88 & 0.3 & 0.39 & 0.94 \\
0.4 & 0.36 & 0.91 & 0.4 & 0.35 & 0.93 \\
\textbf{0.6} & \textbf{0.33} & \textbf{0.92} & \textbf{0.5} & \textbf{0.33} & \textbf{0.92} \\
0.8 & 0.34 & 0.93 & 0.7 & 0.36 & 0.89 \\
1.0 & 0.38 & 0.90 & 0.9 & 0.50 & 0.81 \\
\bottomrule
\end{tabular}%
}
\end{table}

\subsection{Generalization and Efficiency}

\paragraph{Cross-Model Generalization}

We test zero-shot transfer by applying VisuoAlign trained on LLaVA-1.5-7B to InstructBLIP~\cite{dai2023instructblip} and LLaVA-NeXT~\cite{li2024llava}. As shown in Table~\ref{tab:generalization}, it reduces JSR by up to 52\%, indicating that VisuoAlign captures universal safety principles rather than overfitting, and generalizes well across architectures.

\begin{table}[h]
\centering
\caption{Zero-shot generalization on MultiHarm. The alignment from VisuoAlign transfers effectively to new models.}
\label{tab:generalization}
\resizebox{\columnwidth}{!}{%
\begin{tabular}{l cc}
\toprule
\textbf{Model} & \textbf{Base (JSR$\downarrow$/ASR$\uparrow$)} & \textbf{VisuoAlign Transferred (JSR$\downarrow$/ASR$\uparrow$)} \\
\midrule
InstructBLIP & 0.80 / 0.20 & \textbf{0.42 / 0.85} \\
LLaVA-NeXT & 0.76 / 0.24 & \textbf{0.38 / 0.90} \\
\bottomrule
\end{tabular}%
}
\end{table}

\paragraph{Efficiency Analysis}

Although MCTS adds some overhead, Table~\ref{tab:efficiency} shows VisuoAlign increases latency by only ~1.2s per query while keeping memory comparable to other methods. This modest cost is justified by its large safety gains, confirming its practicality for real-world deployment.

\begin{table}[h]
\centering
\caption{Efficiency analysis on LLaVA-1.5-7B. Latency measured as seconds per query.}
\label{tab:efficiency}
\resizebox{0.9\columnwidth}{!}{%
\begin{tabular}{lcc}
\toprule
\textbf{Method} & \textbf{Latency (s/query)}$\downarrow$ & \textbf{Peak Memory (GB)}$\downarrow$ \\
\midrule
Base & 0.31 & 14.5 \\
RLHF & 0.33 & 14.8 \\
DPO & 0.32 & 14.7 \\
\textbf{VisuoAlign} & 1.52 & 15.1 \\
\bottomrule
\end{tabular}%
}
\end{table}
\section{Conclusion}

In this work, we introduced \textbf{VisuoAlign}, a method for multi-modal safety alignment via prompt-guided tree search. By embedding safety constraints into visual–textual interactive prompts, systematically constructing diverse safety-critical trajectories through Monte Carlo Tree Search, and applying prompt-based safety scaling during inference, our approach transforms safety alignment from a static filtering paradigm into a dynamic, reasoning-integrated process. Extensive analysis demonstrates that VisuoAlign enables proactive risk exposure, comprehensive dataset generation, and robust real-time safety adaptation, significantly improving the resilience of LVLMs against cross-modal threats.

\clearpage
\printbibliography


\end{document}